\documentclass[10pt,twocolumn,letterpaper]{article}

\usepackage[pagenumbers]{iccv}      

\definecolor{iccvblue}{rgb}{0.21,0.49,0.74}
\usepackage[pagebackref,breaklinks,colorlinks,allcolors=iccvblue]{hyperref}
\usepackage{multirow}
\usepackage{algorithm}      
\usepackage{algpseudocode}  
\usepackage{pifont}    
\usepackage{booktabs}  
\usepackage{booktabs, adjustbox, multirow}
\usepackage[font=footnotesize]{caption} 
\usepackage{tabularx, makecell, booktabs}
\newcolumntype{C}{>{\centering\arraybackslash}X}
\newcolumntype{R}{>{\raggedleft\arraybackslash}X}
\usepackage{amsmath}
\usepackage{tablefootnote}
\DeclareMathOperator*{\argmax}{arg\,max} 
\DeclareMathOperator*{\argmin}{arg\,min}

\definecolor{iccvblue}{rgb}{0.21,0.49,0.74}
\usepackage[pagebackref,breaklinks,colorlinks,allcolors=iccvblue]{hyperref}

\definecolor{colorCycleGAN}{RGB}{227, 235, 249} 
\definecolor{colorTurbo}{RGB}{255, 250, 224}    
\definecolor{colorVAR}{RGB}{229, 245, 229}      
\definecolor{colorOtherMethods}{RGB}{240, 240, 240} 
\definecolor{MyYellow}{RGB}{184, 134, 11}
\definecolor{MyGreen}{RGB}{127, 210, 127}
\definecolor{Cultured}{HTML}{F2F2F2}
\definecolor{Platinum}{HTML}{DBDBDB}
\definecolor{Spanish Gray}{HTML}{9A9A9A}
\definecolor{Dim Gray}{HTML}{656565}
\definecolor{Onyx}{HTML}{3F3F3F}

\definecolor{Azure X 11 Web Color}{HTML}{EBFEFF}
\definecolor{Aqua}{HTML}{2AF5FF}
\definecolor{Capri}{HTML}{28C2FF}
\definecolor{Blue Jeans}{HTML}{60AFFF}
\definecolor{Blue Pigment}{HTML}{3333AB}
\definecolor{Picotee Blue}{HTML}{262681}
\definecolor{Oxford Blue}{HTML}{002147}

\definecolor{Light Coral}{HTML}{F47983}
\definecolor{Lime Green}{HTML}{32CD32}
\definecolor{Royal Blue (Web Color)}{HTML}{4169E1}

\def\paperID{5599} 
\def\confName{ICCV}
\def\confYear{2025}

\title{CycleVAR: Repurposing Autoregressive Model for Unsupervised One-Step Image Translation }

\author{Yi Liu\\
Beihang University\\ 
{\tt\small 18373214@buaa.edu.cn }
\and
Shengqian Li\\
University of Chinese Academy of Sciences\\ 
{\tt\small 20231111@buaa.edu.cn}
\and
Zuzeng Lin\\
Tianjin University\\ 
{\tt\small linzuzeng@tju.edu.cn}
\and
Feng Wang\\
CreateAI\\
{\tt\small feng.wff@gmail.com}
\and
Si Liu\thanks{Corresponding author}\\
Beihang University\\ 
{\tt\small liusi@buaa.edu.cn}
}

\begin{document}
\maketitle
\begin{abstract}
The current conditional autoregressive image generation methods have shown promising results, yet their potential remains largely unexplored in the practical unsupervised image translation domain, which operates without explicit cross-domain correspondences.
A critical limitation stems from the discrete quantization inherent in traditional Vector Quantization-based frameworks, which disrupts gradient flow between the Variational Autoencoder decoder and causal Transformer, impeding end-to-end optimization during adversarial training in image space.
To tackle this issue, we propose using Softmax Relaxed Quantization, a novel approach that reformulates codebook selection as a continuous probability mixing process via Softmax, thereby preserving gradient propagation. 
Building upon this differentiable foundation, we introduce CycleVAR, 
which reformulates image-to-image translation as image-conditional visual autoregressive
generation by injecting multi-scale source image tokens as contextual prompts, analogous to prefix-based conditioning in language models.
CycleVAR exploits two modes to generate the target image tokens, including (1) serial multi-step generation, enabling iterative refinement across scales, and (2) parallel one-step generation synthesizing all resolution outputs in a single forward pass.
Experimental findings indicate that the parallel one-step generation mode attains superior translation quality with quicker inference speed than the serial multi-step mode in unsupervised scenarios.
Furthermore, both quantitative
and qualitative results indicate that CycleVAR surpasses previous state-of-the-art
unsupervised image translation models, \textit{e}.\textit{g}., CycleGAN-Turbo.

\end{abstract}
\begin{figure}[htbp]
    \centering
    \includegraphics[width=1\linewidth]{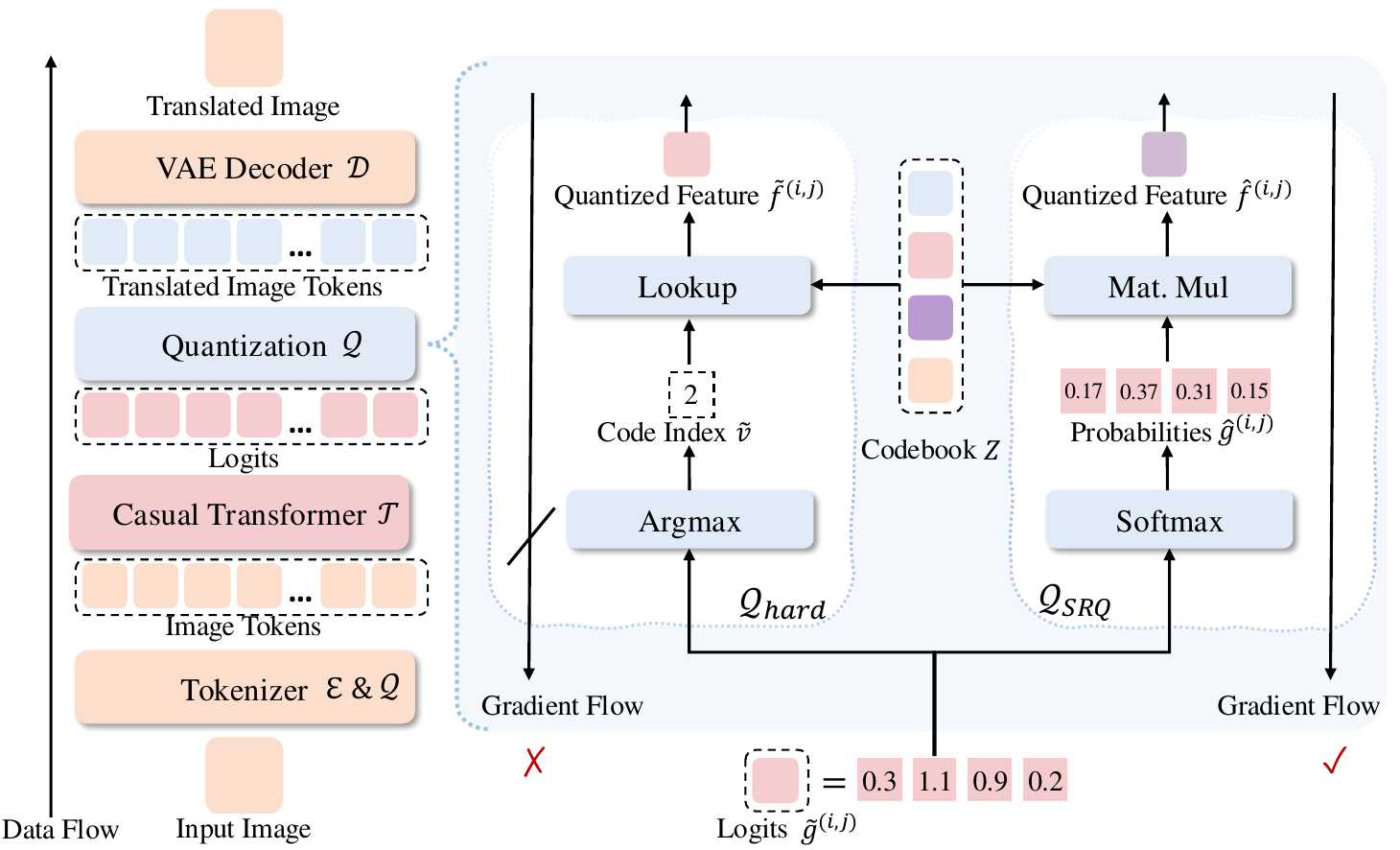}
    \caption{Comparing conventional quantization methods with our Softmax Relaxed Quantization (SRQ) approach in unsupervised image translation. Overcoming gradient backpropagation challenges by replacing non-differentiable argmax with softmax for smooth gradient propagation in SRQ. We assume that the codebook size is $4$ for this picture.}
    \label{fig:motivation}
\end{figure}    
\section{Introduction}
\label{sec:intro}

Recent advancements in image generation have seen explosive growth, particularly in the areas of text-to-image synthesis~\cite{rombach2022high,llamagen,infinity, dalle, sdxl,liu2024playground} and image-to-image translation~\cite{ruiz2023dreambooth,controlnet,controlnet++,t2iadapter}. 
Diffusion models have become the predominant choice in this field due to their outstanding generation quality, open-source ecosystem, and user-friendly training stability.
However, diffusion models entail iterative denoising steps, typically ranging from $20$ to $500$, to generate clear images, resulting in considerable computation cost and inference time.
Furthermore, the denoising-based architecture fundamentally diverges from the autoregressive paradigm of large language models (LLMs)~\cite{touvron2023llama}, hindering integration into unified multimodal models~\cite{emu3}.

These limitations have renewed interest in visual autoregressive models~\cite{parti,llamagen,infinity,emu3,hart,ma2024star,varclip}, which align naturally with LLMs through shared sequential prediction mechanisms. 
Recent studies have delved into employing visual autoregressive models for image-to-image translation through supervised training~\cite{li2024controlvar, li2024controlar}. 
It is relatively straightforward, as we can directly supervise the output logits using the target image token index.
This approach, however, requires extensive paired datasets, which can be impractical for many applications.
Creating aligned pairs for specialized artistic styles entails meticulous effort by professional artists, who must carefully reference original images to produce corresponding new artworks. 
Similarly, translating night scenes to day scenes necessitates capturing identical scenes under strictly controlled conditions. 
The demanding requirements for collecting high-quality paired data can often make the process economically prohibitive or, in some cases, physically infeasible.


The absence of unsupervised image translation methods based on autoregressive models can be attributed to various factors.
First, visual autoregressive models have been less explored and utilized than diffusion models due to the lack of robust open-source implementations. Second, gradient truncation in VQ-based quantization ~\cite{vqvae} disrupts gradient propagation, preventing end-to-end training.
Third, the traditional ``next-token" autoregressive decoding for image synthesis requires sequential steps, which leads to relatively long decoding times and limits practical usability.

In this paper, we propose to utilize \textbf{Softmax Relaxed Quantization (SRQ)}, which replaces discrete vector codebook selection with a continuous probability mixing using Softmax, enabling direct adversarial training in image space and mitigating gradient truncation, as shown in \Cref{fig:motivation}.
Building on this differentiable foundation, we introduce \textbf{CycleVAR}, a novel unsupervised image translation framework that is based on an open-sourced autoregressive model VAR ~\cite{VAR}, which follows coarse-to-fine ``next-scale"
prediction paradigm.
CycleVAR reformulates image-to-image translation as image-conditional visual autoregressive generation by injecting multi-scale tokens tokenized from the source domain image $I$ as contextual prompts, analogous to prefix-based conditioning in language models.
CycleVAR exploits different modes to generate the target image tokens, including (1) serial multi-step generation, enabling iterative refinement across scales, and (2) parallel one-step generation synthesizing all-resolution outputs in a single forward pass.
Experimental results indicate that the parallel one-step mode, in contrast to the serial multi-step mode resembling the original VAR inference process, substantially enhances translation quality while achieving faster inference speed. 
This validates the effectiveness of our SRQ mechanism and multi-scale token prefilling strategy in enabling efficient and high-fidelity synthesis.

Quantitative and qualitative experiments show that CycleVAR outperforms previous state-of-the-art methods utilizing established diffusion models in achieving a better balance between distribution matching and preserving input structure for unpaired translation tasks.

In summary, the main contributions of this article can be summarized as follows:
\begin{itemize}
    \item We introduce Softmax Relaxed Quantization to enable end-to-end training, which replaces codebook sampling with a continuous probability mixing using Softmax. 
    \item We propose the first unsupervised framework, CycleVAR, based on a pre-trained autoregressive image generation model, achieving unpaired image translation through multi-scale image token prefilling and prediction. The proposed CycleVAR further explores two generation modes: serial multi-step generation and parallel single-step generation.
    \item CycleVAR easily expands visual autoregressive models with strong unsupervised generation capability. Under various datasets, the proposed CycleVAR demonstrates its highly competitive performance towards image quality and structure consistency compared to state-of-the-art diffusion methods.
\end{itemize}

\section{Related Works}
\subsection{Autoregressive Image Generation}
In image generation, early autoregressive approaches like~\cite{pixelrcnn, van2016conditional} directly modeled pixel-level dependencies through raster-scan prediction of individual pixel values.
Modern autoregressive image models employ vector quantization (VQ) to convert spatial image patches into discrete tokens, drawing inspiration from NLP's word tokenization.
Through architectures like VQ-VAE~\cite{vqvae}, VQ-VAE-2~\cite{vqvae-2}, RQ-VAE~\cite{rqvae}, and VQ-GAN~\cite{vqgan}, these models achieve efficient autoregressive generation by predicting token indices in latent space.
Recent advancements in this area have led to the development of innovative models. MaskGiT~\cite{maskgit}, for instance, 
generate images by predicting randomly masked tokens.
MAR~\cite{mar} combines diffusion processes with autoregressive generation.
Visual AutoRegressive modeling (VAR)~\cite{VAR} redefines autoregressive image modeling as a coarse-to-fine ``next-scale" prediction framework, significantly enhancing both generation quality and inference speed.

In the realm of text-to-image generation, models such as Parti~\cite{parti}, Open-MAGVIT2~\cite{magvit}, Emu3~\cite{emu3}, HART~\cite{hart}, LlamaGen~\cite{llamagen}, Lumina-mGPT~\cite{liu2024lumina}, and Infinity~\cite{infinity} leverage large-scale data and model sizes to synthesize high-quality images successfully. 
Among them, Infinity is an extension of VAR.
However, there have been limited endeavors to tailor these models for image-conditional generation tasks.

\subsection{Image-to-Image Translation}
Image-to-image translation maps images from a source domain to a target domain. Based on the availability of paired training data, methods are broadly categorized into supervised and unsupervised paradigms.

\textbf{Supervised Image-to-Image Translation} learns mappings between paired domains using annotated datasets. 
The foundational Pix2Pix~\cite{pix2pix} is established on conditional GANs.
Subsequent works incorporate additional constraints~\cite{pix2pixhd, park2019semantic, sangkloy2017scribbler,zhu2020sean}.
Following breakthroughs in text-to-image diffusion models, recent methods~\cite{ruiz2023dreambooth, t2iadapter, controlnet, unicontrolnet, controlnet++, li2023gligen} adapt pre-trained latent diffusion models for controllable translation.
Most recently, ControlAR~\cite{li2024controlar} and ControlVAR~\cite{li2024controlvar} bridge autoregressive image generation (LlamaGen~\cite{llamagen}, VAR~\cite{VAR}) with conditional image control.
However, despite these advances, current supervised approaches remain fundamentally constrained by their reliance on large-scale paired training data. It's a critical limitation in real-world scenarios where pixel-aligned cross-domain pairs are expensive or infeasible to acquire.

\textbf{Unsupervised Domain Translation} addresses the challenge of learning cross-domain mappings without paired supervision. Pioneering works like CycleGAN~\cite{cyclegan}, DualGAN~\cite{yi2017dualgan}, and DiscoGAN~\cite{kim2017learning} established cycle consistency constraints.
Subsequent advances further introduced contrastive learning, enhanced loss, and disentangled representation\cite{cut,han2021dual,shrivastava2017learning,taigman2016unsupervised, munit, lee2018diverse}.
Recent diffusion-based approaches achieved image translation through denoising processes~\cite{wu2023latent, su2022dual,brooks2023instructpix2pix}.
CycleGAN-Turbo~\cite{img2img-turbo} applied cycle consistency loss with distilled model SD-Turbo ~\cite{sauer2024adversarial} to achieve one-step translation. 
Despite these innovations, current methods exhibit a critical limitation: no existing framework successfully integrates unsupervised translation with pre-trained visual autoregressive models, leaving their potential for cross-domain generalization unexplored. 
\section{Method}
\begin{figure*}[th!]
    \centering
    \includegraphics[width=0.95\linewidth]{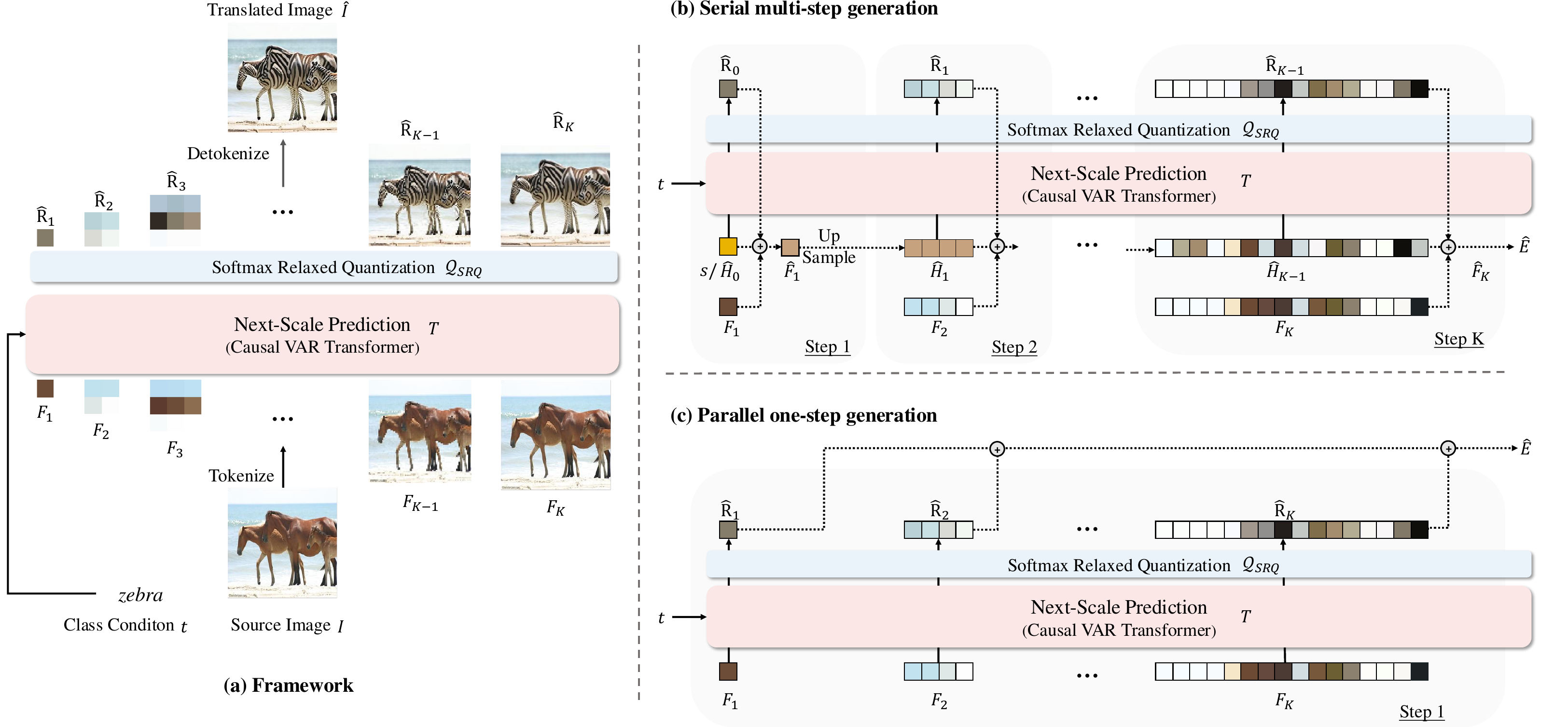}
    \caption{Illustration of CycleVAR. (a) The framework of repurposing VAR for image translation by prefilling multi-scale image tokens and making multi-scale prediction; (b) Serial multi-step generation mode following the ``step-by-step" paradigm; (c) Parallel one-step generation mode by concurrently feeding all scale image tokens. $\hat{E}$ will be delivered to the VAE decoder $\mathcal{D}$ to decode the translated image $\hat{I}$}
    \vspace{-0.2in}
\label{fig:framework}
\end{figure*}
\subsection{Preliminary}
Our method is based on VAR architecture~\cite{VAR}, which incorporates a discrete visual tokenizer and a causal VAR Transformer for image synthesis with condition $t$\footnote{For the class-conditional image generation model VAR, $t$ is the class label. For the text-conditional model Infininty, $t$ is the input text prompt. Our subsequent descriptions of our model structure are based on VAR.}.
The discrete visual tokenizer includes a Variational Auto-Encoder(VAE) and a discrete vector quantizer. The vector quantizer contains a learnable codebook $Z\in \mathbb{R}^{V\times C}$ with $V$ vectors.

\noindent{}\textbf{Tokenization.} 
The VAE encoder $\mathcal{E}(\cdot)$ first encodes the ground truth image $I \in \mathbb{R}^{H\times W \times 3}$ in $E \in \mathbb{R}^{h\times w \times C}$, then the vector quantizer quantifies the feature map $E$ into multiscale residual maps $R_1, R_2, ...R_K$, where $R_k \in \mathbb{R}^{h_k\times w_k\times C}$ and $h_k, w_k$ grows gradually as $k$ ranges from $1$ to $K$.
During the quantization process, each feature vector $f^{(i,j)}$ obtained from $E$ is mapped to the discrete token code index $q^{(i, j)}$ of its nearest code in the Euclidean sense as follows:
\begin{align}
    q^{(i,j)} = \left( \argmin_{v \in [V]} \| \text{lookup}(Z, v) - f^{(i,j)} \|_2 \right) \in [V],
\end{align}
where $\text{lookup}(Z, v)$ means taking the $v$-th vector in $Z$. 
Then the VAE decoder $\mathcal{D}(\cdot)$ reconstructed $\tilde{I}$ with $\tilde{E}$ and $\tilde{E}$ is comprised of every $\text{lookup}(Z, q^{(i,j)})$.
The training of the tokenizer is accomplished through a combination of the discriminator and reconstruction losses.

\noindent{}\textbf{Causal VAR Transformer.} 
Upon pre-training the visual tokenizer, the causal VAR Transformer is subsequently trained to predict residuals
$R_K$ conditioned on previous $R_1, R_2, ... R_{K-1}$ and $t$, the autoregressive likelihood is as follows:
\begin{align}
\label{eq:serial}
    p(R_1, R_2, ..., R_K)  = \prod_{k=1}^{K} p(R_k | R_1, ..., R_{k-1}, t).
\end{align}
An index-wise classifier maps the output feature of the last layer of transformer $\tilde{h}^{(i, j)}$ to logits $\tilde{g}^{(i, j)} \in \mathbb{R}^{V}$ among the entire codebook. 
The logits are supervised using cross-entropy loss with the corresponding label $y^{(i,j)} \in [0, V)$.

During prediction, the quantizer assigns the index with max logits as the predicted codebook index $\tilde{v}$ and gets the predicted quantized feature $\tilde{f}^{(i, j)}$ as follows:
\begin{align}
    \tilde{f}^{(i,j)} &= \mathcal{Q}_{hard}(\tilde{g}^{(i, j)}) = \text{lookup}(Z, \tilde{v})\\
    &=\text{lookup}(Z, \argmax_{v \in [V]}(\tilde{g}^{(i, j)})).
\end{align}

Conditioned with the class embedding $s$ as the start token, the causal VAR Transformer produces
$\tilde{R}_1, \tilde{R}_2, ... \tilde{R}_{K}$ step by step and uses decoder $\mathcal{D}(\cdot)$ to generate the predicted class-conditional image.

\subsection{CycleVAR}



We propose \textbf{CycleVAR}, which reformulates image-to-image translation as image-conditional visual autoregressive generation by injecting multi-scale tokens tokenized from the source image $I$ as contextual prompts, analogous to prefix-based conditioning in language models as shown in \Cref{fig:framework}(a). 
We freeze the visual tokenizer's parameters during fine-tuning and tokenize\footnote{We employ the straight-through estimator to ensure gradient propagation, enabling the incorporation of cycle loss in training.} the source image $I$ into residual maps $R_k$ of different scales.
By adding the multi-scale residual maps $R_k$ together, we get the features $F_k$:
\begin{align}
    F_k = \text{down}(\sum_{k=1}^{K}\text{up}(R_k, (h, w)), (h_k, w_k)),
\end{align}
where $\text{up}(\cdot)$ and $\text{down}(\cdot)$ represent upsampling and downsampling operators. These $F_k$ will be the context input to the causal VAR Transformer.
Based on the prefilling concept, we introduce two innovative generation modes.
\subsubsection{Serial Multi-Step Generation}
Inspired by the ``step-by-step" thought of
autoregressive generation, we construct outputs through $K$ refinement steps sequentially as shown in \Cref{fig:framework}(b), each step conducts:
\begin{align}
    \hat{H}_k &=  \text{up}\Bigl(\bigl(a \cdot(\hat{R}_{k-1} + \hat{H}_{k-1}) + (1-a)\cdot F_k\bigr), (h_k,w_k)\Bigr),
\end{align}
where $\hat{H}_0$ is initialized with the class embedding $s$.
$a$ is the weight assigned to the fusion of the source and the generated feature.
$\hat{R}_k$ is obtained from the $k$-th forward:
\begin{align}
    \hat{R}_k &= \mathcal{Q}\Bigl(\mathcal{T}\bigl(\hat{H}_{k},\underbrace{(\hat{H}_0,\ldots,\hat{H}_k)}_{\text{Contextual Keys}},t\bigr)\Bigr),
\end{align}
where $\mathcal{Q}$ is the Quantization operation. $\mathcal{T}(q,k,t)$ represents the causal VAR Transformer, which conducts attention with queries generated from $q$ and contextual keys from $k$.
$t$ is the condition of AdaLN used in the casual Transformer. 
Final output $\hat{E}$ is equal to $\hat{H}_{K}$.
We can get the translated image $\hat{I} = \mathcal{D}(\hat{E})$.
Serial multi-step generation requires $K$ forward passes per iteration.

\subsubsection{Parallel One-Step Generation}
Instead of feeding different scales $F_k$ to the causal VAR Transformer progressively, we propose to process all scales concurrently through:
\begin{align}
    \hat{R}_k &= \mathcal{Q}\Bigl(\mathcal{T}\bigl(F_k,\underbrace{(F_1,\ldots,F_k)}_{\text{Contextual Keys}},t\bigr)\Bigr).
\end{align}
In this way, we can get all $R_k$ simultaneously as shown in \Cref{fig:framework}(c).
The final output $\hat{E}$ will be obtained by progressive upsampling and summation across scales:
\begin{align}
    \hat{E} &= a \cdot \sum_{k=1}^K \text{up}\bigl(\hat{R}_k,(h_K,w_K)\bigr) + (1-a) \cdot F_K.
\end{align}
Parallel one-step generation requires only one model forward pass, resulting in faster training and inference speed.

\subsection{Softmax Relaxed Quantization}
Our objective is to learn a mapping from source domain images $I \sim \mathcal{S}$ to target domain images $\hat{I} \sim \mathcal{T}$ under unsupervised settings where paired training data is unavailable. This poses two fundamental challenges:
\begin{enumerate}[leftmargin=*,label=(\roman*)]
    \item The lack of ground truth target images prevents direct supervision of codebook index $\hat{y}^{(i,j)}$.
    \item The non-differentiable $\argmax(\cdot)$ operator in conventional vector quantization creates gradient flow discontinuity during end-to-end optimization.
\end{enumerate}

To address these challenges, we propose using \textbf{Softmax Relaxed Quantization (SRQ)}, which reparameterizes codebook vector selection through continuous probability mixing. The SRQ framework consists of two core components.
We first get soft probabilities $\hat{g}_v^{(i,j)}$ for each codebook vector $Z_v$ by applying softmax to logits $\tilde{g}^{(i, j)}$:
\begin{align}
    \hat{g}^{(i,j)}_v &= \frac{\exp(\frac{\tilde{g}_v^{(i,j)}}{\tau})}{\sum_{q=1}^V \exp(\frac{ \tilde{g}_q^{(i,j)}}{\tau})}, \quad \tau > 0 \label{eq:srq} 
\end{align}
where $\tau$ is the temperature that controls distribution sharpness.
Subsequently, we apply a weighted average to all codebook vectors based on the relaxed distribution,
\begin{align}
    \hat{f}^{(i,j)} &= \mathcal{Q}_{\text{SRQ}}(\tilde{g}^{(i,j)}) = \sum_{v=1}^V \hat{g}^{(i,j)}_v Z_v. \label{eq:srq_average}
\end{align}
$\tau \to 0$ recovers one-hot encoding, which is approximate to the $\argmax(\cdot)$ operation.


\subsection{Training Losses}
We follow CycleGAN-Turbo~\cite{img2img-turbo} to conduct unpaired training between unpaired datasets $\mathcal{X}$ and $\mathcal{Y}$.
For every image from source domain $I_x \in \mathbb{R}^{H\times W \times 3}$, we will randomly sample an image $I_y \in \mathbb{R}^{H\times W \times 3} $from the target domain.
Our framework implements bidirectional cross-domain translation through dual mapping functions:
\begin{equation}
    \mathcal{G}(I_x, t_{y}): \mathcal{X} \rightarrow \mathcal{Y} \quad \text{and} \quad \mathcal{G}(I_y, t_{x}): \mathcal{Y} \rightarrow \mathcal{X}.
\end{equation}

Both transformations employ the unified network architecture $\mathcal{G}$, and $t_x, t_y$ represents the condition of the target area.
The training is composed of cycle consistency loss $\mathcal{L}_{cycle}$, adversarial loss $\mathcal{L}_{gan}$, and identity regularization loss $\mathcal{L}_{idt}$ for generator $\mathcal{G}$.
The discriminator $\mathcal{C}$ is training with corresponding adversarial loss $\mathcal{L}_{dis}$.
The cycle consistency loss will encourage the generated image $\hat{I}$ to be translated back to its source image $I$.
The adversarial loss encourages the generated image $\hat{I}$ to be similar to the target domain.
The discriminator $\mathcal{C}_x, \mathcal{C}_y$ is trained to distinguish between real and generated images.

The cycle consistency loss will encourage the generated image $\hat{I}$ to be translated back to its source image $I$ as follows:
\begin{equation}
    \begin{aligned} 
        \mathcal{L}_{cycle} & = \mathbb{E}_x[\mathcal{G}(\mathcal{G}(I_x, t_y), t_x), I_x] \\
             & + \mathbb{E}_y[\mathcal{G}(\mathcal{G}(I_y, t_x), t_y), I_y].
    \end{aligned}
\end{equation}
The adversarial loss encourages the generated image $\hat{I}$ to be similar to the target domain.
The discriminator $\mathcal{C}_x, \mathcal{C}_y$ is trained to distinguish between real and generated images.
For the generator $\mathcal{G}$, the adversarial loss is:
\begin{equation}
    \begin{aligned} 
        \mathcal{L}_{gan} & = \mathbb{E}_x[log(\mathcal{C}_y(\mathcal{G}(I_x, t_y)))] + \mathbb{E}_y[log(\mathcal{C}_x(\mathcal{G}(I_y, t_x)))].
    \end{aligned}
\end{equation}
For the discriminator $\mathcal{C}$, the adversarial loss is:
\begin{equation}
    \begin{aligned} 
        \mathcal{L}_{dis} & = \mathbb{E}_x[log\mathcal{C}_x(I_x)] + \mathbb{E}_x[log(1-\mathcal{C}_y(\mathcal{G}(I_x, t_y)))] \\
        &+ \mathbb{E}_y[log\mathcal{C}_y(I_y)] +  \mathbb{E}_y[log(1-\mathcal{C}_x(\mathcal{G}(I_y, t_x)))].
    \end{aligned}
\end{equation}
The identity loss is:
\begin{equation}
    \begin{aligned} 
        \mathcal{L}_{idt} & =  \mathbb{E}_x[\mathcal{L}_{rec}(\mathcal{G}(I_x, t_x), I_x)] + \mathbb{E}_y[\mathcal{L}_{rec}(\mathcal{G}(I_y, t_y), I_y)],
    \end{aligned}
\end{equation}
where the $\mathcal{L}_{rec}$ is composed of L1 loss and LPIPS difference loss. 
\begin{table*}[t] 
  \centering
\resizebox{0.95\textwidth}{!}{%
\begin{tabular}{@{}ccccc|cccc@{}}
\toprule
\multirow{2}{*}{Method} & \multicolumn{2}{c}{Horse $\rightarrow$ Zebra}                                            & \multicolumn{2}{c|}{Zebra $\rightarrow$ Horse}                                           & \multicolumn{2}{c}{Day $\rightarrow$ Night}                                              & \multicolumn{2}{c}{Night $\rightarrow$ Day}                                              \\ \cmidrule(l){2-9} 
                        & FID $\downarrow$ & \begin{tabular}[c]{@{}c@{}}DINO\\ Structure $\downarrow$\end{tabular} & FID $\downarrow$ & \begin{tabular}[c]{@{}c@{}}DINO\\ Structure $\downarrow$\end{tabular} & FID $\downarrow$ & \begin{tabular}[c]{@{}c@{}}DINO\\ Structure $\downarrow$\end{tabular} & FID $\downarrow$ & \begin{tabular}[c]{@{}c@{}}DINO\\ Structure $\downarrow$\end{tabular} \\ \midrule
CycleGAN~\cite{cyclegan}                & 74.9             & 3.2                                                                   & 133.8            & 2.6                                                                   & 36.3             & 3.6                                                                   & 92.3             & 4.9                                                                   \\
CUT~\cite{cut}                     & 43.9             & 6.6                                                                   & 186.7            & 2.5                                                                   & 40.7             & 3.5                                                                   & 98.5             & 3.8                                                                   \\ \midrule
SDEdit~\cite{meng2021sdedit}                  & 77.2             & 4.0                                                                   & 198.5            & 4.6                                                                   & 111.7            & 3.4                                                                   & 116.1            & 4.1                                                                   \\
Plug\&Play~\cite{plugplay}              & 57.3             & 5.2                                                                   & 152.4            & 3.8                                                                   & 80.8             & 2.9                                                                   & 121.3            & \textbf{2.8}                                                          \\
Pix2Pix-Zero~\cite{parmar2023zero}            & 81.5             & 8.0                                                                   & 147.4            & 7.8                                                                   & 81.3             & 4.7                                                                   & 188.6            & 5.8                                                                   \\
Cycle-Diffusion~\cite{wu2023latent}         & \textbf{38.6}    & 6.0                                                                   & 132.5            & 5.8                                                                   & 101.1            & 3.1                                                                   & 110.7            & 3.7                                                                   \\
DDIB~\cite{su2022dual}                    & 44.4             & 13.1                                                                  & 163.3            & 11.1                                                                  & 172.6            & 9.1                                                                   & 190.5            & 7.8                                                                   \\
InstructPix2Pix~\cite{brooks2023instructpix2pix}         & 51.0             & 6.8                                                                   & 141.5            & 7.0                                                                   & 80.7             & \textbf{2.1}                                                          & 89.4             & 6.2                                                                   \\ \midrule
CycleGAN-Turbo~\cite{img2img-turbo}          & 41.0             & 2.1                                                                   & 127.5            & \textbf{1.8}                                                          & {\underline{31.3}}       & {\underline{3.0}}                                                             & {\underline{45.2}}       & 3.8                                                                   \\ \midrule
CycleVAR w/ VAR          & 47.8             & \textbf{1.8}                                                         & {\underline{124.1}}      & {\underline{1.9}}                                                            & -\tablefootnote{As ImageNet does not contain specific classes for day or night, we did not conduct day $\leftrightarrow$ night experiments on CycleVAR w/ VAR(a class-conditional image generation model pre-trained on ImageNet).}                 & -                                                                    & -                & -                                                                     \\
CycleVAR w/ Infinity    & {\underline{40.3}}       & {\underline{1.9}}                                                             & \textbf{119}     & \textbf{1.8}                                                          & \textbf{30.6}    & {\underline{ 3.0}}                                                             & \textbf{43.6}    & {\underline{3.2}}                                                             \\ \bottomrule
\end{tabular}%
}
\caption{Comparison results on $256 \times 256$ and $512 \times 512$ datasets. The best results are highlighted in bold, and the second-best results are underlined.
We adopt the parallel one-step generation mode for this comparison.}
\label{tab:main}
\end{table*}

\section{Experiment}
\begin{figure*}[ht]
    \centering
\includegraphics[width=0.95\linewidth]{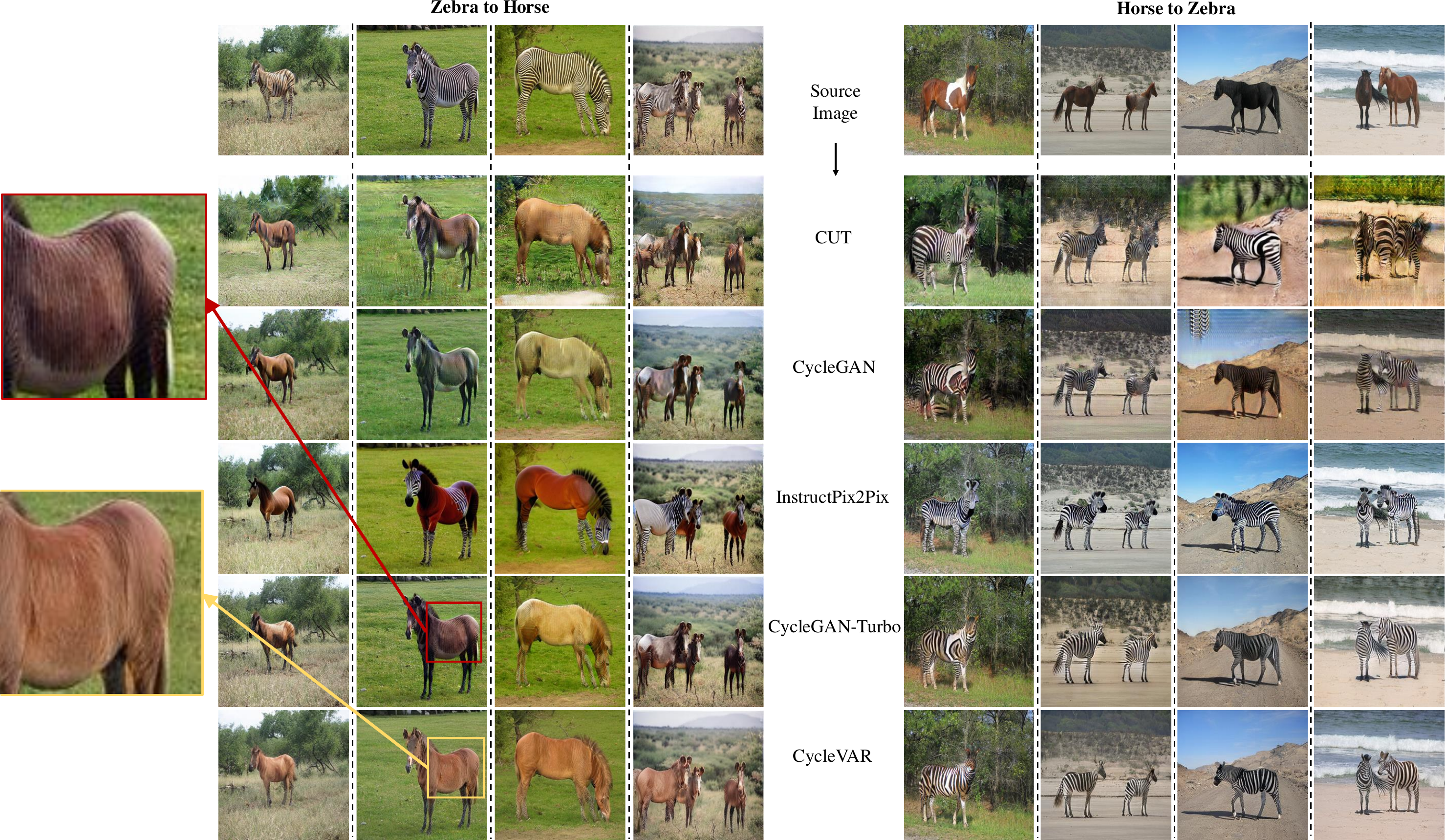}
    \caption{
    Qualitative comparison on horse $\leftrightarrow$ zebra dataset($256 \times 256$). Our method, CycleVAR, is contrasted with GAN-based and diffusion methods. CycleVAR demonstrates strong performance in image translation while maintaining the integrity of the source image structure. Additionally, it ensures consistent transfer across foreground objects while preserving the chromatic harmony of the background.
    }
    \vspace{-0.2in}
\label{fig:h2z}
\end{figure*}
\subsection{Settings}
\noindent \textbf{Dataset.}
We perform unpaired translation experiments on two widely utilized datasets with varying resolutions(horse $\leftrightarrow$ zebra, day $\leftrightarrow$ night from BDD100k~\cite{yu2020bdd100k}).
For the horse $\leftrightarrow$ zebra dataset, we follow CycleGAN~\cite{cyclegan} to load
$286\times286$ images and use random $256\times256$ crops when training.
During inference,
we apply translation at $256 \times 256$. 
For the day $\leftrightarrow$ night dataset, we resize all images to $512 \times 512$ during training and inference. 
For evaluation, we use
the corresponding validation sets for horse $\leftrightarrow$ zebra and select $500$ images for day $\leftrightarrow$ night following CycleGAN-turbo~\cite{img2img-turbo}.
We also conduct experiments on the Anime Scene Dataset~\cite{jiang2023scenimefy}, which contains real scene photos from  Landscapes High-Quality (LHQ)
dataset~\cite{lhq} and anime scene photos from nine prominent Shinkai
Mokoto films, to show better visualization results.
We apply translation to the Anime Scene dataset at $256 \times 256$.

\noindent \textbf{Training details.}
We use the VAR-310M~\cite{VAR} model to conduct the ablation study with $256\times 256$ resolutions on the horse $\leftrightarrow$ zebra dataset.
Moreover, we use the text-conditional image generation model, Infinity-2B~\cite{infinity}, an extension of vanilla VAR, to conduct qualitative and quantitative comparative experiments in parallel one-step mode.
The discriminator uses the CLIP model as the feature extractor and an MLP as the decoder.
We employ the AdamW optimizer with a constant learning rate, incorporating a warmup phase to stabilize training.
More details on training are provided in the supplementary material.

\noindent \textbf{Evaluation metric.}
We use the Frechet Inception Distance(FID) to calculate the distribution matching between generated and target area images.
A lower FID score represents better alignment with the target distribution, indicating greater authenticity.
Following CycleGAN-turbo~\cite{img2img-turbo}, we use DINO-Structure-Dist to measure the structure similarity between the generated and source area image.
We report DINO-Structure-Dist multiplied by $100$
, where a lower DINO-Struct-Dist reflects improved retention of the input structure within the translated image.
In subsequent text, we abbreviate DINO-Struct-Dist as DINO Structure for brevity

\subsection{Main results}
\begin{table*}
\centering
\vspace{-0.2in}
\begin{tabular}{@{}cccccc@{}}
\toprule
Method  & AnimeGANv3~\cite{chen2020animegan} & White-box~\cite{whitebox} & CartoonGAN~\cite{chen2018cartoongan} & Scenimefy~\cite{jiang2023scenimefy} & CycleVAR       \\ \midrule
Style   & 0.035      & 0.078     & 0.080      & 0.157     & \textbf{0.650} \\
Content        & 0.257      & 0.067     & 0.100      & 0.165     & \textbf{0.411} \\
Overall        & 0.063      & 0.046     & 0.111      & 0.171     & \textbf{0.609} \\ \bottomrule
\end{tabular}
\caption{User preference study in Anime Scene Dataset with shinkai style. The optimal scores are highlighted in bold. We compare one-step CycleVAR w/ Infinity with other state-of-the-art baselines designed for image translation in anime. The best results are in bold.}
\label{tab:human}
\end{table*}
\begin{figure*}
    \centering
    \includegraphics[width=0.9\linewidth]{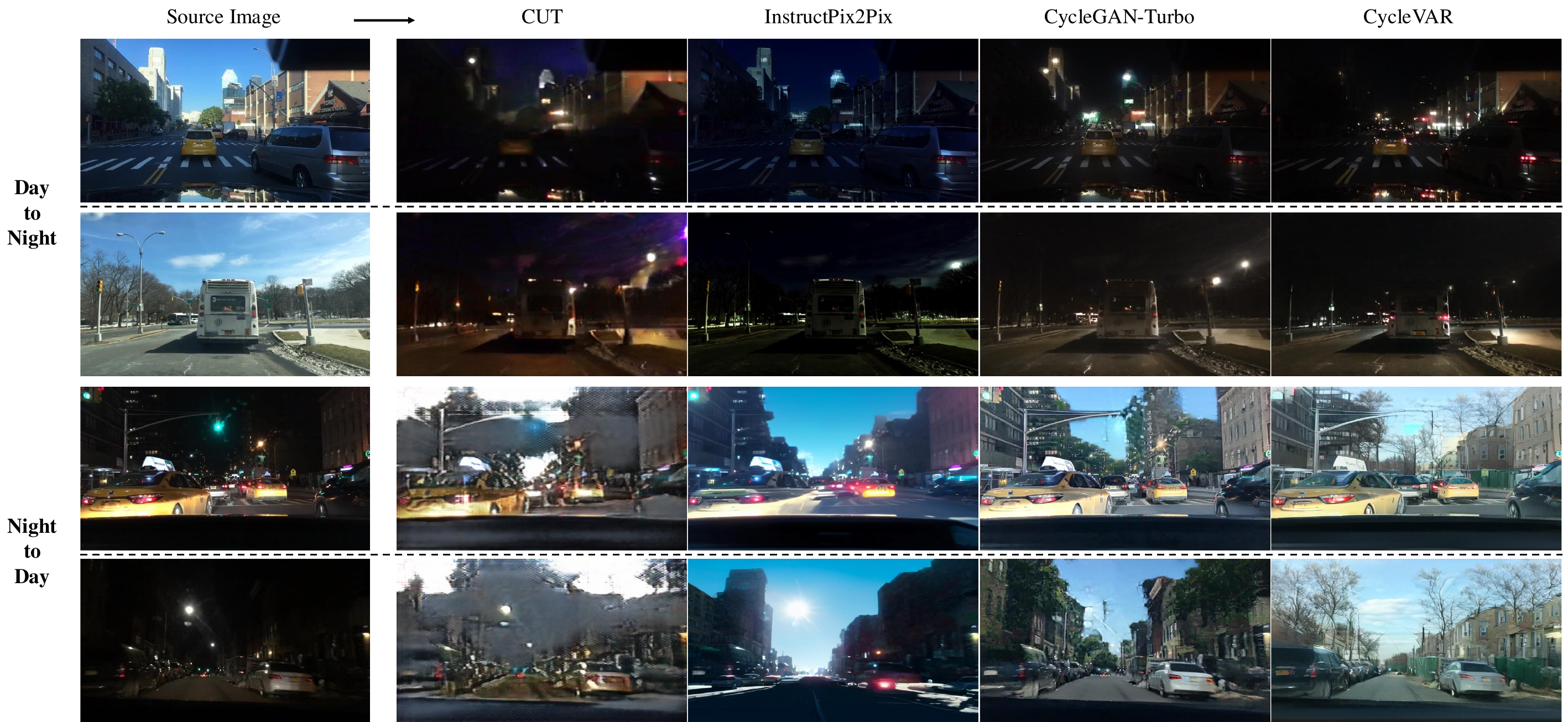}
    \caption{Qualitative comparison on day$\leftrightarrow$night dataset in the driving scene.}
    \vspace{-0.2in}
\label{fig:d2n}
\end{figure*}
\subsubsection{Quantitative Comparison}
We first compare CycleVAR to state-of-the-art GAN-based unpaired image translation methods, zero-shot image editing techniques, and diffusion models tailored for image editing in standard CycleGAN datasets.
As shown in \Cref{tab:main}, CycleVAR achieves an optimal balance between visual realism and structural integrity, whereas existing methods frequently sacrifice one aspect for the other.

Compared to traditional GAN-based methods, our approach exhibits clear superiority in FID and DINO Structure metrics across various datasets, as shown in the first two rows of \Cref{tab:main}.
Furthermore, when compared with zero-shot image translation methods as shown in rows $3$rd to $7$th of \Cref{tab:main}, CycleVAR w/ VAR-310M can achieve the desired balance between FID and DINO Structure Score.
It should be emphasized that all these zero-shot methods are based on a pre-trained text-to-image diffusion model with extensive training data, while CycleVAR w/ VAR employs VAR-310M, which is trained exclusively on ImageNet~\cite{deng2009imagenet}.
For a broader comparison, we also evaluate our method against InstructPix2Pix, a general-purpose model for text-based image editing that fails to preserve the original image structure.
The most related work with us is CycleGAN-Turbo, which applies adversarial loss to a pre-trained text-to-image diffusion model, Stable Diffusion Turbo (v2.1).
Stable Diffusion Turbo is distilled from Stable Diffusion to achieve one-step inference.
Conversely, our method directly fine-tunes the visual autoregressive generation model to achieve one-step image translation, getting better results.

Regarding the horse $\leftrightarrow$ zebra and day $\leftrightarrow$ night dataset may not adequately reflect the aesthetic of the generated results, we conduct a user preference study on the Anime Scene Dataset to assess the quality of generation from three perspectives following Scenimefy~\cite{jiang2023scenimefy}:
1. Style: evident anime stylization. 2. Content: consistent structure and semantic preservation. 3. Overall: overall translation performance.
For the user preference study in Table \ref{tab:human}, ten sets of images are shuffled and anonymously distributed to participants, who will select the best result from five methods in each set according to different criteria.
We summarized the results from 46 participants to calculate the average preference scores.
As shown in \Cref{tab:human}, our method receives the best scores in all three criteria. 
Our method surpasses Scenimefy, a semi-supervised technique that generates pseudo labels using StyleGAN~\cite{stylegan1, stylegan2}, providing compelling evidence of our effectiveness.

\begin{figure}
    \centering
    \includegraphics[width=1\linewidth]{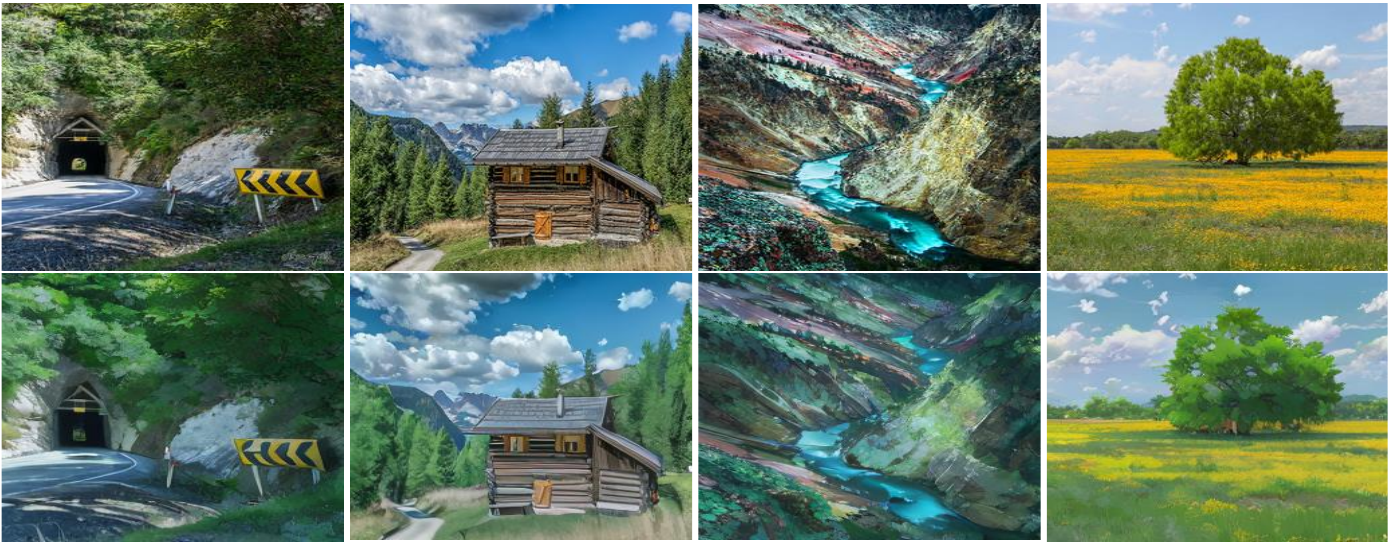}
    \vspace{-0.2in}
    \caption{Anime Scene Translation via parallel one-step CycleVAR w/ Infinity. (Top) Input domain: natural scene photographs. (Bottom) Output domain: CycleVAR-generated images emulating Shinkai's signature style.}
    \vspace{-0.2in}
    \label{fig:shinkai}
\end{figure}

\subsubsection{Qualitative Results}
\Cref{fig:h2z} shows the qualitative comparison with the object-centric dataset horse $\leftrightarrow$ zebra.
For zebra $\rightarrow$ horse translation, InstructPixPix only generates part of a realistic zebra's appearance.
While other methods may seem to deliver acceptable translation results at first glance, a closer examination reveals that the zebra's distinctive stripe pattern remains unchanged, with only a modification in color intensity.
Our approach, however, successfully transforms the zebra's stripes into a fur texture.
This achievement is likely attributed to integrating multi-scale context and output in our methodology.
Similarly, our approach has achieved a more realistic effect in the horse $\rightarrow$ zebra transformation.

For the harder day$\leftrightarrow$ night dataset in the driving scene, except for the overall color, daytime photos feature a light blue sky and white clouds with no distinct lighting or metallic sheen. 
In contrast, nighttime photos depict a pitch-black sky with noticeable lighting and reflective effects, as shown in the first row of Figure $\ref{fig:d2n}$.
CUT struggles to generate a normal picture.
InstructPix2Pix and CycleGAN-Turbo can change the overall color, but they can not produce the above nighttime and daytime features.
In contrast, the images produced by CycleVAR present a dark night sky, noticeable lighting, and reflection of light in the car for day $\rightarrow$ night.
Meanwhile, CycleVAR can generate light blue sky and white clouds and diminish the scattered bright spots in the image for night $\rightarrow$ day.

We also showcase results on an Anime Scene Dataset, as shown in Figure \ref{fig:shinkai}.
More visual comparisons between CycleVAR and anime scene translated state-of-the-art methods are provided in the supplementary material.

\subsection{Ablation Study}


\begin{table}[htb]

\begin{adjustbox}{max width=\columnwidth, center}
\begin{tabular}{cccccc}
\hline
\multirow{2}{*}{Generation Mode} & \multirow{2}{*}{\begin{tabular}[c]{@{}c@{}}Inference \\ Time\end{tabular}} & \multicolumn{2}{c}{Horse $\rightarrow$ Zebra}                                              & \multicolumn{2}{c}{Zebra $\rightarrow$ Horse}                                               \\ \cline{3-6} 
                                 &                                                                            & FID $\downarrow$ & \begin{tabular}[c]{@{}c@{}}DINO \\ Structure  $\downarrow$\end{tabular} & FID  $\downarrow$ & \begin{tabular}[c]{@{}c@{}}DINO\\  Structure  $\downarrow$\end{tabular} \\ \hline
Serial Multi-Step                & 0.22                                                                       & 82.05            & 1.90                                                                    & 189.80            & \textbf{1.80}                                                           \\
Parallel One-Step                & \textbf{0.08}                                                              & \textbf{47.82}   & \textbf{1.77}                                                           & \textbf{124.10}   & 1.85                                                                    \\ \hline
\end{tabular}
\end{adjustbox}
\caption{Ablation of generation mode for CycleVAR. The unit of measurement for inference time is seconds(s).}
\label{tab:main_abl}
\end{table}
\noindent\textbf{Generation Mode.}
We propose two novel generation modes based on the prefilling concept for image-to-image translation: serial multi-step generation and parallel one-step generation.
The results corresponding to these modes are presented in \Cref{tab:main_abl}. 
The inference time shown in Table \ref{tab:main_abl} was measured on the Nvidia A100 80G GPU, with a single input image size of $256\times256$.
For each generation mode, we conducted $10$ measurements and calculated the average.
The inference time encompasses the processing time through the causal VAR Transformer and Softmax Relaxed Quantization. 
In the serial multi-step mode, we employed KV Cache~\cite{kvcache} technology for acceleration.
It was observed that parallel one-step generation can achieve comparable structural preservation outcomes to serial generation while better facilitating the transformation of the image domain. 
On the contrary, serial multi-step generation faces obstacles in efficiently converting the source image to the target domain while also exhibiting prolonged inference time.
The calculation details of inference time are proved in the supplementary material.
These subpar translated results could be attributed to the repeated application of SRQ in the serial multi-step process. In an unsupervised environment, the absence of individual supervision for each step leads to the accumulation of errors from SRQ over $K$ iterations, resulting in suboptimal optimization outcomes.

\begin{table}[htb]
\begin{adjustbox}{max width=\columnwidth, center}
\begin{tabular}{cccccc}
\hline
M-S                   & M-S                   & \multicolumn{2}{c}{Horse $\rightarrow$ Zebra}                                             & \multicolumn{2}{c}{Zebra $\rightarrow$ Horse}                                             \\ \cline{3-6} 
Context              & Output               & FID $\downarrow$ & \begin{tabular}[c]{@{}c@{}}DINO \\ Structure $\downarrow$\end{tabular} & FID $\downarrow$ & \begin{tabular}[c]{@{}c@{}}DINO\\  Structure $\downarrow$\end{tabular} \\ \hline
$\checkmark$         & $\checkmark$         & \textbf{47.82}   & \textbf{1.77}                                                          & \textbf{124.10}  & \textbf{1.85}                                                          \\
$\checkmark$         &                      & 58.70            & 2.10                                                                   & 130.60           & 2.28                                                                   \\
\multicolumn{1}{l}{} & \multicolumn{1}{l}{} & 268.70           & 4.39                                                                   & 260.00           & 5.41                                                                   \\ \hline
\end{tabular}
\end{adjustbox}

\caption{Ablation study for multi-scale impact in one-step parallel generation. "M-S" stands for "multi-scale". The best results are in bold.}

\label{tab:parallel}
\end{table}
\begin{table}[htb]
\begin{adjustbox}{max width=1\columnwidth, center}
\begin{tabular}{@{}ccccc@{}}
\toprule
\multirow{2}{*}{Temperature} & \multicolumn{2}{c}{Horse $\rightarrow$ Zebra}                                            & \multicolumn{2}{c}{Zebra $\rightarrow$ Horse}                                            \\ \cmidrule(l){2-5} 
                             & FID $\downarrow$ & \begin{tabular}[c]{@{}c@{}}DINO\\ Structure $\downarrow$\end{tabular} & FID $\downarrow$ & \begin{tabular}[c]{@{}c@{}}DINO\\ Structure $\downarrow$\end{tabular} \\ \midrule
0.01                         & 228.20           & \textbf{1.23}                                                                  & 231.20           & \textbf{1.35}                                                                  \\
0.10                         & 60.67            & 1.85                                                                  & 138.70           & 1.98                                                                  \\
0.70                         & 48.59            & 1.82                                                                  & 127.50           & 1.91                                                         \\
1.00                         & 48.39            & 1.80                                                                  & 125.30           & 1.88                                                                  \\
2.00                         & 47.82            & 1.77                                                         & 124.10           & 1.85                                                                  \\
10.00                        & \textbf{47.11}   & 1.88                                                                  & \textbf{120.00}  & 1.98                                                                  \\
10000.00                     & 457.90           & 1.33                                                                  & 487.20           & 1.40                                                                  \\
0.70 wo/ gumbel              & 48.58            & 1.77                                                        & 127.40           & 1.92                                                                  \\ \bottomrule
\end{tabular}
\end{adjustbox}
\caption{Ablation of the temperature in Softmax Relaxed Quantization for one-step parallel generation. The best results are in bold.}
\vspace{-0.1in}
\label{tab:temperature}
\end{table}

\noindent\textbf{Multi-Scale Context and Output.}
We analyze the impact of multi-scale image tokens as context and multi-scale image tokens as output in the parallel one-step generation by progressively dropping them.
When solely incorporating $\hat{R}_K$ with $F_K$ in the final output, while omitting $\hat{R}_1,..., \hat{R}_{K-1}$, suboptimal outcomes are obtained. Yet, it is still possible to generate relatively normal translated results.
However, the accuracy decreased significantly when we further masked $R_1,..., R_{K-1}$ for $R_K$ during the self-attention operation. It can only generate images with large color blocks, vaguely outlining the objects.
It shows that the different scale image tokens contain richer information, which benefits the image translation.

\begin{figure}
    \centering
    \includegraphics[width=1.0\linewidth]{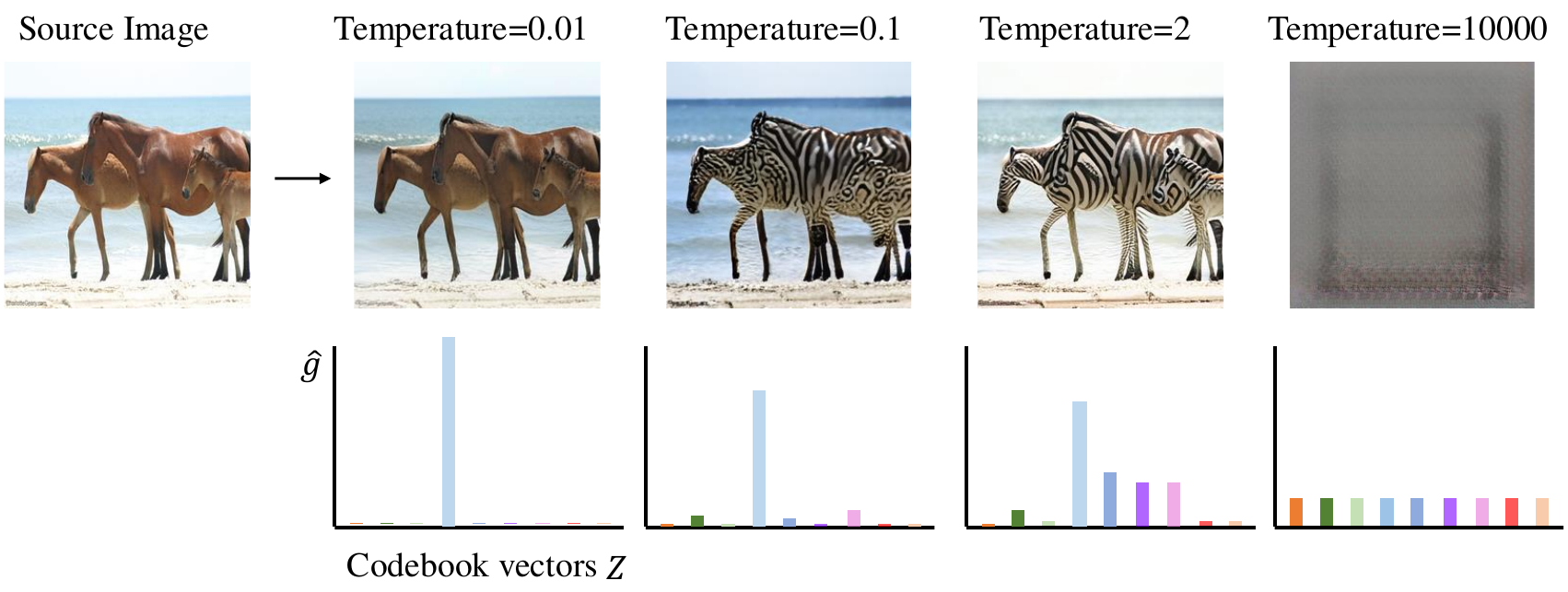}
    \vspace{-0.2in}
    \caption{Visual ablation of the temperature in Softmax Relaxed Quantization for parallel one-step generation.}
    \vspace{-0.2in}
\label{fig:temperature}
\end{figure}

\noindent\textbf{Temperature in Softmax Relaxed Quantization.}
The smaller temperature $\tau$ will make a sharper distribution more like a one-hot distribution, as shown in the second column of \Cref{fig:temperature}.    
However, the sharp distribution does not facilitate gradient propagation, resulting in the inability to transfer the image.
Therefore, the DINO Structure is best in the $1$st row of Table \ref{tab:temperature}.
The translated result gets better with softer distribution as the temperature increases.
However, extremely high temperatures (\textit{e}.\textit{g}., $\tau = 10,000$) induce a uniform distribution, causing outputs to collapse to the mean codebook vector and lose semantic meaning, as shown in the last column in \Cref{fig:temperature}. 
We initially inject Gumbel-distributed noise into the SRQ (\Cref{eq:srq}) to induce controlled stochasticity during training. Surprisingly, results in \Cref{tab:temperature} (bottom row) reveal that removing this noise component yields negligible performance variance ($\Delta\text{FID} < 0.1$), suggesting the model remains stable under deterministic quantization.

\section{Conclusion}
In this study, we analyze the challenge of repurposing the visual autoregressive model for unsupervised image translation, specifically focusing on how discrete quantization disrupts gradient flow. 
We introduce Softmax Relaxed Quantization to maintain gradient propagation and CycleVAR to redefine image-to-image translation as image-conditional visual autoregressive generation with prefilling multi-scale image tokens. 
Additionally, we propose two generation modes: the serial multi-step mode and the parallel one-step mode.
Our experiments highlight the superiority of the parallel one-step mode for CycleVAR under the unsupervised setting. 
Experiments across diverse datasets demonstrate that CycleVAR is capable of delivering high-quality results and conserving the original image's structural information, surpassing the other contemporary methods.


{
    \small
    \bibliographystyle{ieeenat_fullname}
    \bibliography{main}
}

\maketitlesupplementary 
\begin{figure*}[htb!]
    \centering
    \includegraphics[width=0.9\linewidth]{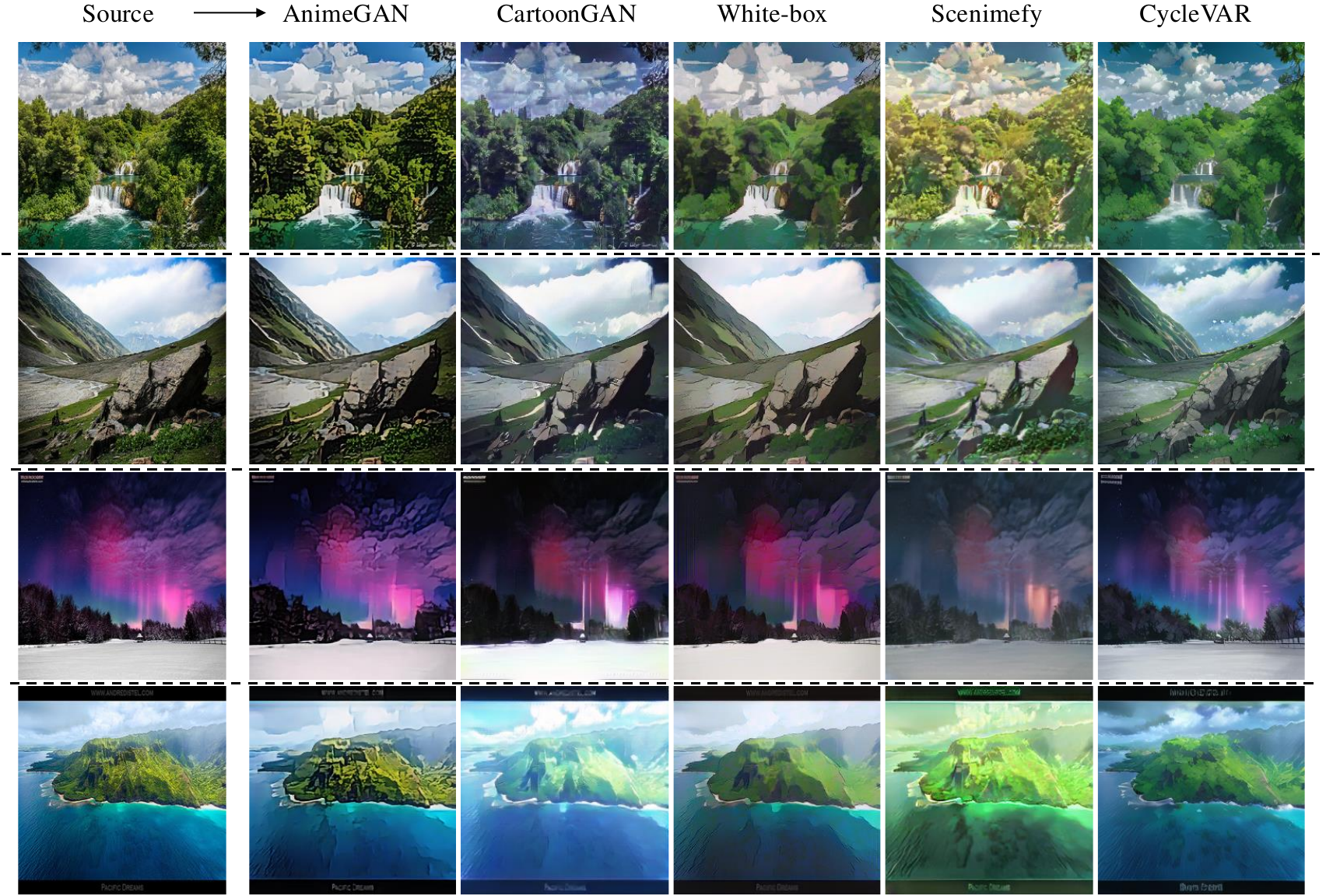}
    \vspace{-0.1in}
    \caption{Qualitative comparison on Anime Scene Dataset. We compare parallel one-step CycleVAR w/ Infinity with other state-of-the-art baselines designed for image translation in anime. }
    \vspace{-0.2in}
\label{fig:shinkai_compare}
\end{figure*}
\setcounter{section}{0}
\section{Multi-Step Serial Generation}
\begin{algorithm}[H]
\caption{Multi-Step Serial Generation}
\small 
\begin{algorithmic}[1]
\Require Source features $\{F_k\}_{k=1}^K$, class condition $t$, class embedding of $s$
\Ensure Final output $\hat{H}_K$
\State Initialize $\hat{H}_0 \gets s $
\For{$k = 1$ \textbf{to} $K$}
    \State $\tilde{G}_{k-1} \gets \text{VARTransformer}(\hat{H}_{k-1}, (\hat{H}_0,\ldots,\hat{H}_{k-1}), t)$ 
    \State $\hat{R}_{k-1} \gets \text{Quantization}(\tilde{G}_{k-1})$ 
    \State $\hat{F}_{k} \gets a\cdot (\hat{R}_{k-1} + \hat{H}_{k-1}) + (1-a)\cdot F_k$
    \If{$k < K$}
             \State $\hat{H}_{k} \gets \text{Upsample}(\hat{F}_{k}, (h_{k+1},w_{k+1}))$ 
    \Else
            \State $\hat{H}_{k} \gets \hat{F}_{k}$ 
    \EndIf
    
\EndFor

\State \Return $\hat{H}_K$
\end{algorithmic}
\label{algo:serial}
\end{algorithm}
The complete process of Multi-Step Serial Generation is shown in Algorithm \ref{algo:serial}.
\section{CycleVAR w/ Infinity}
Instead of using the original vector quantizer of vanilla VAR, Infinity uses a dimension-independent bitwise quantizer, which is implemented by LFQ and BSQ.
The input feature $f^{(i, j)}$ is quantized to $q^{(i, j)}$ as follows:
\begin{equation}
q^{(i, j)} = \mathcal{Q}_{sign}(f^{(i, j)}) =
\begin{cases} 
\mathrm{sign}(f^{(i, j)}) & \text{if } \mathrm{LFQ} \\
\frac{1}{\sqrt d} \mathrm{sign}(\frac{f^{(i, j)}}{|f^{(i, j)}|}) & \text{if } \mathrm{BSQ}
\end{cases}
\label{eq:lfq}
\end{equation}
where $\mathrm{sign}(\cdot)$ is the piecewise function that extracts the sign of a real number.
$V=2^d$ for the bitwise quantizer. 
Instead of using the index-wise classifier like VAR, the casual VAR transformer of Infinity predicts features' labels with $d$ binary classifiers in parallel to predict whether each dimension of $\tilde{f}^{(i,j)}$ is positive or negative.

When implementing Softmax Relaxed Quantization with the bitwise quantizer, a comparable method is employed: The non-differentiable sampling operation based on logits after the classifier is substituted with softmax applied to the prediction of each binary classifier.

\section{Training Details}
We employed two models with the ``next-scale" prediction paradigm as baselines: the class-conditional image generation model VAR~\cite{VAR} and the text-conditional generation model Infinity~\cite{infinity}, with Infinity being an extension of VAR.

\noindent\textbf{CycleVAR w/ VAR.} 
When translating horse to zebra, we assign the class label of zebra as $t$ for input to the casual VAR Transformer. In contrast, the casual VAR Transformer incorporates the horse class label as condition $t$.

\noindent\textbf{CycleVAR w/ Infinity.}
In CycleVAR w/ Infinity, the text prompt serves as the condition $t$. The associated text in the horse $\leftrightarrow$ zebra dataset is ``picture of a horse" and ``picture of a zebra." Likewise, in the day $\leftrightarrow$ night dataset, the text is ``driving in the day" and ``driving in the night." As for the Anime Scene Dataset, the text corresponds to ``a Makoto Shinkai style landscape" and ``a photo of a real landscape."
\section{Inference Time} 
\begin{table}[htbp]
  \centering
  \caption{Inference Time Comparison. The unit of time is seconds.}
  \label{tab:inference}
  \footnotesize
  \setlength{\tabcolsep}{9pt} 
  \addtolength{\extrarowheight}{1.5pt} 

  \begin{tabular}{cc  cc}
    \toprule
    Method & Time & Method & Time  \\ 
    \cmidrule(r{0.5em}l{0.5em}){1-2} \cmidrule(r{0.5em}l{0.5em}){3-4}

    \cellcolor{colorCycleGAN}CycleGAN            & \cellcolor{colorCycleGAN}0.004 & \cellcolor{colorOtherMethods}SDEdit              & \cellcolor{colorOtherMethods}1.900    \\
    \cellcolor{colorCycleGAN}CUT                 & \cellcolor{colorCycleGAN}0.004 & \cellcolor{colorOtherMethods}Plug\&Play          & \cellcolor{colorOtherMethods}6.300   \\
    \cellcolor{colorCycleGAN}                    & \cellcolor{colorCycleGAN}      & \cellcolor{colorOtherMethods}Pix2pix-Zero        & \cellcolor{colorOtherMethods}14.20  \\
    \cellcolor{colorTurbo}CycleGAN-Turbo      & \cellcolor{colorTurbo}0.080 & \cellcolor{colorOtherMethods}Cycle-Diffusion     & \cellcolor{colorOtherMethods}3.500  \\
    \cellcolor{colorVAR}CycleVAR w/ VAR     & \cellcolor{colorVAR}0.030 & \cellcolor{colorOtherMethods}DDIB                & \cellcolor{colorOtherMethods}3.900  \\
    \cellcolor{colorVAR}CycleVAR w/ Infinity& \cellcolor{colorVAR}0.110 & \cellcolor{colorOtherMethods}InstructPix2Pix     & \cellcolor{colorOtherMethods}4.200   \\
    \bottomrule
  \end{tabular}
\end{table}

We also present a comparative analysis of inference times across various methods, with all evaluations conducted on an Nvidia H100 80G GPU. The results are summarized in Table~\ref{tab:inference}.
Traditional GAN-based approaches (\textcolor{Blue Jeans}{blue rows}) are characterized by high inference speeds and small model sizes, with approximately 12~Million active parameters. 
However, this efficiency is typically achieved at the expense of lower generation quality. In contrast, methods based on non-distilled Stable Diffusion (\textcolor{Dim Gray}{gray rows}) exhibit substantially slower performance due to their iterative denoising process. 
These models are larger; for instance, Stable Diffusion 1.4 and 1.5 are built upon a 1.1~Billion-parameter architecture, while Stable Diffusion 2.1 utilizes a 1.3~Billion-parameter model.
CycleGAN-Turbo (\textcolor{MyYellow}{yellow row}) achieves one-step inference by applying LoRA to a distilled SD-Turbo 2.1 model, utilizing 1.1~Billion relevant parameters. 
Our proposed method, CycleVAR (\textcolor{MyGreen}{green rows}), demonstrates a more favorable balance between efficiency and performance. By directly fine-tuning original autoregressive models, CycleVAR achieves rapid inference using the 420~Million- parameter VAR-310M and maintains comparable speed with the larger Infinity-2B, which has 2.3~Billion parameters for causal VAR Transformer. 
This highlights CycleVAR's ability to deliver high-quality results while remaining computationally efficient, representing a significant advancement over existing trade-offs between speed and fidelity.

\section{Qualitative and Quantitative Results}
\label{sec:shinkai_compare}

Our method produces clearer images with richer texture details while capturing the light and dark color characteristics reminiscent of Shinkai's style, as shown in Figure \ref{fig:shinkai_compare}.
The images generated by AnimeGAN focus too much on the edges, resulting in excessive contrast and a lack of Shinkai's distinctive style.
The overall color scheme of CartoonGAN's images tends to be either too washed out or too dark.
Conversely, White-box-generated images exhibit basic cartoon color characteristics, but the details are blurred and lack finer granularity.
Scenimefy preserves more details than White-box; however, the images still appear blurry and lack the delicate presentation of light spots, failing to evoke a romantic and fantastical atmosphere.


\end{document}